\documentclass[sigconf]{acmart}

\usepackage{amsmath,amsfonts}
\usepackage{siunitx}
\usepackage{array}
\usepackage[caption=false,font=normalsize,labelfont=sf,textfont=sf]{subfig}
\usepackage{textcomp}
\usepackage{stfloats}
\usepackage{url}
\usepackage{verbatim}
\usepackage{graphicx}
\usepackage{lipsum}
\usepackage{balance}

\usepackage{wasysym}
\usepackage[justification=centering]{caption}
\usepackage{subfig} 
\usepackage{subfloat}
\usepackage{multirow}
\usepackage{xcolor}
\usepackage{diagbox}
\graphicspath{{./figures/}}
\usepackage{array}
\usepackage[ruled,noline,noend]{algorithm2e}
\usepackage{flushend}

\newtheorem{Definition}{Definition}

\newtheorem{theorem}{\textbf{Theorem}}

\begin{document}

\setcopyright{acmlicensed}
\copyrightyear{2018}
\acmYear{2018}
\acmDOI{XXXXXXX.XXXXXXX}

\acmYear{2025}
\acmConference[Preprint]{Proceedings of Preprint}{}{}
\acmBooktitle{Proceedings of Preprint}

\title{Lossless Privacy-Preserving Aggregation for Decentralized Federated Learning}

\author{Xiaoye Miao$^{\ast\star}$, Bin Li$^{\ast}$, Yanzhang$^{\ast}$, Xinkui Zhao$^{\ddagger}$, Yangyang Wu$^{\ddagger}$}
\affiliation{\institution{$^\ast$Center for Data Science, Zhejiang University, Hangzhou, China}
\institution{$^\ddagger$College of Software, Zhejiang University, Ningbo, China}
\institution{$^\dagger$College of Computer Science, Zhejiang University, Hangzhou, China} \country{}}
\affiliation{\{lb0, miaoxy, nameyzhang, zhaoxinkui, wuyy \}@zju.edu.cn \country{}}
\begin{abstract}
Privacy concerns arise as sensitive data proliferate.
Despite \emph{decentralized federated learning} (DFL) aggregating gradients from neighbors to avoid direct data transmission, it still poses indirect data leaks from the transmitted gradients.
Existing differential privacy (DP) methods for DFL add random noise to
gradients. They not only diminish the model accuracy but also suffer from ineffective gradient protection.
In this paper, we propose a \emph{novel lossless privacy-preserving aggregation} rule named \textsf{LPPA} to enhance gradient protection as much as possible but without losing DFL model accuracy.
\textsf{LPPA} subtly injects the \emph{noise difference} between the sent noise and received noise into transmitted gradients for gradient protection. This noise difference incorporates neighbors’ randomness for each client, effectively safeguarding against data leaks.
Moreover, \textsf{LPPA} employs the noise flow conservation theory to ensure that all noise impact can be globally eliminated. The global sum of all noise differences remains zero, ensuring that accurate global gradient aggregation is unaffected and the model accuracy remains intact. We theoretically prove that the privacy-preserving capacity of \textsf{LPPA} is $\sqrt{2}$ times greater than that of DP, while maintaining comparable model accuracy to the standard DFL aggregation without noise injection.
Experimental results verify the theoretical findings and show that \textsf{LPPA} achieves a 14\% mean improvement in accuracy over the DP method. 
We also demonstrate the effectiveness of \textsf{LPPA} in protecting raw data and guaranteeing lossless model accuracy.

\end{abstract}





\maketitle

\section{Introduction}
Federated learning (FL) is a prominent paradigm in distributed machine learning, significantly enhancing data utilization across multiple data owners \cite{DBLP:journals/kais/LiuHZLJXD22}.
In FL, multiple clients (i.e., data owners) jointly train a \emph{global} model in an iterative manner, not sharing their local training data with the central server.  
This paradigm is \emph{heavily dependent} on the central server, since the server is responsible for distributing and aggregating the model updates from each client per iteration. 
Once the server is attacked or fails, the entire process is at risk of being compromised \cite{DBLP:journals/pami/KumarMC24}. 
As a consequence, \emph{decentralized} federated learning (DFL) has emerged as an alternative paradigm, in which each client acts as an \emph{independent} aggregation server beyond their original local training role \cite{DBLP:conf/icml/Shi0WS00T23}. 
It is not reliant on the central server, particularly well-suited for applications in edge computing, WoT systems and financial collaboration \cite{10.1145/3637528.3671942,DBLP:conf/www/0001LL0L24,10.1145/3637528.3671582}.



In the DFL setting, each client directly sends its model update to \emph{neighboring} clients, and \emph{autonomously} aggregates the received updates following the \emph{aggregation rule}. Then they can use the aggregated model updates to refine their local model. Different DFL methods essentially use different aggregation rules, such as simply averaging all received updates \cite{DBLP:journals/pami/SunLW23}, which ensures all DFL clients converge to an \emph{optimal and consistent} model (i.e., transitioning from initially disparate local models to a unified global model) \cite{nedic2020distributed}.

However, the client's local training data in DFL can still be leaked from the transmitted model updates (i.e., model weights and gradients). 
In other words, an adversarial client in DFL can receive model gradients from its neighbor, and infer this neighbor's raw training data from the received gradients \cite{DBLP:journals/network/YangGXXLL24}. 
For instance, for a simple differentiable model, the gradients directly correspond to the multiples of raw data \cite{lyu2018privacy}, thus enabling a straightforward inference of raw data from received gradients. 
For more complex models, the adversarial client can simulate the dummy data and gradients to reversely approximate the true raw data, via minimizing the discrepancy between the dummy gradients and truly received gradients (a.k.a. \emph{data reconstruction attack} \cite{DBLP:journals/tifs/YangGXL23}). 
Thus, without adequate protection, the adversarial client can reconstruct all sender's raw data. It is an invasion unacceptable to other participating clients.


As a result, to enhance gradient protection and raw data security within the DFL aggregation process, several privacy-preserving aggregation are developed. These methods aim to prevent adversary from accessing precise local gradients (i.e., each client's local model gradient), via resource-intensive \emph{cryptographic} schemes or trade-off \emph{differential privacy} methods. 
\emph{First}, cryptographic schemes \cite{guo2021privacy} ensure that, the adversary only receives encrypted local gradients, thereby complicating the attainment of precise gradients without the corresponding key. However, constructing key systems in DFL introduces significantly greater complexity, due to the pairwise key requirements between every two clients.
These schemes also require additional encryption and decryption per iteration, which are not suitable for DFL with limited resources.
\emph{Second}, the differential privacy (DP) methods \cite{truex2020ldp,DBLP:conf/www/0003FFSTXL22} safeguard raw data by adding random noise to local gradients before transmission,  preventing the adversary from obtaining precise gradients and reconstructing full raw data. However, DP method necessitates a \emph{trade-off}  between model accuracy and gradient protection to control the noise addition level. They \emph{not} effectively defend against data leaks and \emph{degrade} the model accuracy \cite{wei2020federated}. 
In addition, in the DP method, each client's local model is \emph{damaged} owing to aggregating the noise-added gradients received from neighbors.
This may lead to reluctance to participate in the DFL, driven by concerns about selfishness \cite{yu2020fairness}.

Therefore, in this paper, we attempt to explore a strategic noise injection method for defending against data leaks from transmitted gradients, while preferably minimizing the DFL model accuracy loss.  By analyzing the noise addition in DP method, we identify \emph{two} principal challenges, associated with using noise to protect sender's local gradients and, consequently, the raw training data.

(i) \emph{How can noise be strategically injected into gradients to provide effective gradient protection?} The traditional DP method assumes that the adversary can only receive gradients \cite{abadi2016deep}. 
However, in the DFL setting, the model structure, data dimension, and aggregation rule are \emph{uniformly} applied across adversarial and normal clients to train the global model.
Given this uniformity, after receiving local gradients derived from sender's raw data, the adversary can easily reconstruct sender's raw data by optimizing the discrepancy between received gradients and its simulated gradients.
Adding random noise to local gradients only \emph{marginally} reduces the integrity of reconstructed data; however, this form of gradient protection is far from effective, as evidenced by our experiments in this work.

(ii) \emph{Can the noise impact on global model accuracy be eliminated?} In the DFL setting, the high global model accuracy heavily depends on the accurate aggregation of local gradients. It essentially enables all local gradients to consistently converge to zero across clients \cite{DBLP:conf/icml/Shi0WS00T23}.
However, the \emph{irreducible} noise added by DP method introduces disordered perturbations across local gradients, which is \emph{persistent} throughout entire aggregation process and gradually amplifies the perturbation impact \cite{sun2021fl}.
Over time, this irreducible noise leads to significant divergences among local gradients, severely impeding consistent convergence and ultimately diminishing model accuracy.

To address above challenges, we propose a \emph{novel lossless privacy-preserving DFL aggregation rule} named \textsf{LPPA} in this paper. Inspired by the flow conservation in graph theory  \cite{chen2012applied}, we find that from a global perspective, the overall noise inflow equals the overall noise outflow within the DFL connected topology.
Therefore, \textsf{LPPA} can employ an extra round of noise exchange (i.e., noise flow) to preemptively confuse DFL clients, thereby rendering clients insensitive to subsequently transmitted gradients. It meanwhile enables each client to inject the noise difference between sent and received noise into their local gradients, rather than pure noise. This noise difference injection not only enhances local gradient protection with stronger randomness, but also facilitates global noise elimination via flow conservation, thereby ensuring \emph{lossless} model accuracy.
Our primary contributions can be described as: 
\begin{itemize}
    \item We introduce a novel aggregation rule named \textsf{LPPA} for DFL, 
    designed to maintain lossless model accuracy after noise injection while providing effective gradient protection.

    \item For gradient protection, a preemptive communication round is employed to exchange random noise and inject the noise difference into local gradients. It incorporates neighbors' randomness for each client to enhance gradient protection.

    \item For model accuracy, as the global sum of all noise differences remains zero per iteration, accurate gradient aggregation and consistent convergence can be steadily guaranteed, thus ensuring the lossless global model accuracy eventually.

   \item We prove lossless accuracy of \textsf{LPPA} in comparison with the standard DFL aggregation without noise injection. We analyze the privacy budget of \textsf{LPPA} and deduce its privacy-preserving capacity is $\sqrt{2}$ times that of DP method. Extensive experiments indicate that, \textsf{LPPA} provides 14\% higher accuracy and more effective protection than DP method.
    
\end{itemize}

The rest of this paper is organized as follows. Section \ref{section-2}  reviews the related work. Section \ref{section-3} introduces the problem definition. Section \ref{section-4} overviews our proposed aggregation rule. Section \ref{section-5} conducts the performance analysis, including the lossless model accuracy and effective gradient protection of \textsf{LPPA}. Section \ref{section-6} presents the experimental results. Section \ref{section-7} concludes this work.

\vspace*{-5pt}
\section{Related Work}
\label{section-2}
\textbf{Decentralized federated learning (DFL)} aggregation rules can be categorized into two types based on the presence or absence of gradient tracking property, i.e., distributed stochastic gradient descent (DSGD) and distributed stochastic gradient tracking (DSGT). DSGD aggregation rules \cite{10.5555/3295222.3295285,9771388} represent a natural extension of stochastic gradient descent (SGD) in decentralized topology, weighted averaging received gradients and utilizing the double-stochastic matrix to guarantee consistent convergence. However, DSGD aggregation underperforms in heterogeneous data partitioning \cite{pmlr-v119-koloskova20a}. DSGT aggregation rules, including \cite{9789732,10.5555/3540261.3541134}, primarily employ an auxiliary variable to track the global gradient (i.e., the average of all local gradients), thus achieving optimal convergence under heterogeneous data partitioning \cite{DBLP:conf/nips/AketiH023}. However, these aggregations all suffer from gradient exposure and immediate data leaks, and we specifically focus on the DSGT aggregation and its protection.

\textbf{Data leaks within DFL} often occur following the gradient transmission in DSGT, as local gradients inherently contain valuable information about the raw training data \cite{lyu2018privacy}. By analyzing received gradients, the adversary can infer the membership of data samples \cite{9109557}, obtain the label distribution \cite{NEURIPS2021_0d924f0e}, and even directly reconstruct the raw training data and labels \cite{DBLP:series/lncs/Zhu020}. For example, in the data reconstruction attack, the adversary first receives the model weight and local gradient from the sender. Subsequently, it simulates the dummy input data and corresponding label, which, along with the received model weight, are fed into the model to back-propagate the dummy gradient. Then adversary can optimize the dummy data to closely approximate the true raw data by minimizing the discrepancy between the dummy gradient and the truly received gradient. Ideally, the simulated dummy data would ultimately match the sender's raw data, that is, reconstructing all raw data. In this work, we consider defending against this data reconstruction attack.


\textbf{Differential privacy (DP) methods} are widely employed to enhance the gradient protection of FL and DFL. These methods primarily add random noise into model gradients, which can reduce the adversary's sensitivity to the received gradients. DP method can be categorized into two types: local, global and distributed. The local DP method \cite{DBLP:conf/ijcai/SunQC21} adds random noise into local gradients before aggregation to ensure individual protection. Conversely, The global DP method \cite{9945997} adds random noise to gradients after aggregation to provide central protection. The distributed DP method \cite{10387777} also adds random noise in distributing setting, i.e., DFL.
However, these methods inevitably compromise the accuracy of the original gradient, eventually leading degrading the global model accuracy. Hence, we aim to investigate a noise injection strategy that minimizes the loss on global model accuracy while providing effective protection.

\section{Problem Definition}
\label{section-3}


In particular, we consider $N$ clients $C=\left\{c_1, c_2, \ldots, c_N\right\}$ participating in DFL aggregation, each client $c_i$ holds a local dataset $\mathcal{D}_i$. For illustrative purposes, we do not make specific distributional assumptions about how training data is sampled across clients. 
For each client $c_i$, $\mathcal{N}^{out}_i$ comprises the set of out-neighbor clients receiving information from $c_i$, and $\mathcal{N}^{in}_i$ comprises the set of in-neighbor clients sending information to $c_i$.
Each client assigns the aggregation weight $w_{ij}$ to its in-neighbor $c_j \in \mathcal{N}^{in}_i$ (including itself), while setting this weight to zero for other non-communicating clients. The aggregation weight matrix $\boldsymbol{W}=[w_{ij}]_{n \times n}$ should be a double-stochastic matrix, i.e., $\sum_{i=1}^N w_{ij}=1$ and $\sum_{i=1}^N w_{li}=1$, which guarantees consistent convergence across model weight \cite{pu2020push}.  
Define $\mathcal{D}= \cup_{i=1}^N \mathcal{D}_i$ as the collective union of all datasets.

\textbf{The objective of DFL} is for all clients to collaboratively train a unified global model while ensuring that each client's local training data remains internal \cite{nedic2020distributed,DBLP:conf/nips/YangZW22}, as stated below. 

\begin{Definition}
Let $\mathcal{L}(\cdot)$ represent the  loss function across all DFL clients, and $\boldsymbol{\theta}_i$ represent the local model weight for client $c_i$. The optimization objective of DFL can be expressed as follows.
\begin{equation}
\label{Definition-2}
        \mathop{\min} \sum_{i=1}^N \mathcal{L}(\boldsymbol{\theta}_i) \quad  s.t.\quad
           \boldsymbol{\theta}_1=\boldsymbol{\theta}_2=\cdots=\boldsymbol{\theta}_N
\end{equation}
\end{Definition}

In this paper, we employ the practical DSGT aggregation rule \cite{10.5555/3540261.3541134}.
It mainly utilizes an \emph{auxiliary variable} for each client to track the global gradient in DFL (i.e., the average of all local gradients). Specifically, in the $t$-th iteration, each client $c_i$ in DSGT aggregation updates its model weight and auxiliary variable as follows.
\begin{align}
        \boldsymbol{\theta}_i^{t+1} &\leftarrow \sum_{j=1}^N w_{ij} \boldsymbol{\theta}_j^{t}-\lambda \boldsymbol{\gamma}_i^t  \label{dsgt-w}\\
        \boldsymbol{\gamma}_i^{t+1} &\leftarrow \sum_{j=1}^N w_{ij}\boldsymbol{\gamma}_j^{t} + \nabla \mathcal{L}(\boldsymbol{\theta}_i^{t+1}) -\nabla \mathcal{L}(\boldsymbol{\theta}_i^{t}) \label{dsgt-y}
\end{align}
Each client primarily aggregates the model weights and auxiliary variables to collaboratively find the optimal model weight $\boldsymbol{\theta}^*$. Moreover, the aggregation of these auxiliary variables allows them to systematically track the global gradient, i.e., $\sum_{i=1}^N \boldsymbol{\gamma}_i^t = \sum_{i=1}^N \nabla \mathcal{L}(\boldsymbol{\theta}_i^{t})$. This implies that auxiliary variables not only tend toward consistency but also constantly track the global gradient, thereby guiding the optimization process more effectively \cite{pu2020push}. 
To establish this global gradient tracking, in the initialization, each client $c_i$ need to set the initial auxiliary variable $\boldsymbol{\gamma}_i^0$ equal to its initial local gradient $\nabla \mathcal{L}(\boldsymbol{\theta}_i^0)$, i.e., $\boldsymbol{\gamma}_i^0=\nabla \mathcal{L}(\boldsymbol{\theta}_i^0)$. 
We will refer to the auxiliary variable $\boldsymbol{\gamma}$ as the \emph{gradient tracking variable} hereafter.

It is worthwhile to point out that, in the DSGT aggregation, \emph{gradient exposure} and \emph{data leaks} typically occur, since the initial gradient tracking variable is completely \emph{equal} to the initial local gradient. Transmitting this variable will directly expose the sender's local gradient and leak its raw data. Thus, it has to pay significant consideration to gradient protection during initialization. 
In subsequent aggregation, the gradient tracking variable gradually evolves from representing the local gradient to reflecting the global gradient,
which also leaks certain data and needs suitable protection.

\textbf{Adversary’s Capabilities and Goals.}
Without loss of generality, in this paper, we consider the most common ``honest but curious'' adversary in data leaks \cite{le2023privacy}. 
In particular, this adversary honestly transmits and aggregates gradient tracking variables following the prescribed aggregation rule.
It possesses the same model structure, computational, and communication capabilities as those of normal clients.
The distinguishing characteristic of this adversary is its intent to infer sender's raw training data by executing data reconstruction attacks on the received gradient tracking variables.

\textbf{Problem statement}. 
The study in this paper attempts to develop an improved DFL aggregation rule on top of DSGT, in order to pursue the \emph{lossless model accuracy} and \emph{enhanced gradient protection}. 
In other words, the predictive accuracy of the global model derived by this newly proposed DFL aggregation rule should be equivalent to that of the global model aggregated by the standard DSGT (that is without noise injection).
At the same time, this new rule should provide competitive gradient protection against data leaks, i.e., effectively prevent the ``honest but curious” adversary from reconstructing sender's raw data from gradient tracking variables. 

In the following discussion, we will refer to the standard DSGT aggregation rule without noise injection simply as standard DSGT, and the DGST aggregation rule augmented with the DP method, which includes adding random noise to local gradients \cite{10387777}, as DP method, to simplify our terminology.

\section{Solution Overview}
\label{section-4}

\begin{figure*}
\centering
 \includegraphics[trim={0.3cm 0cm 0cm 0cm},scale=0.8]{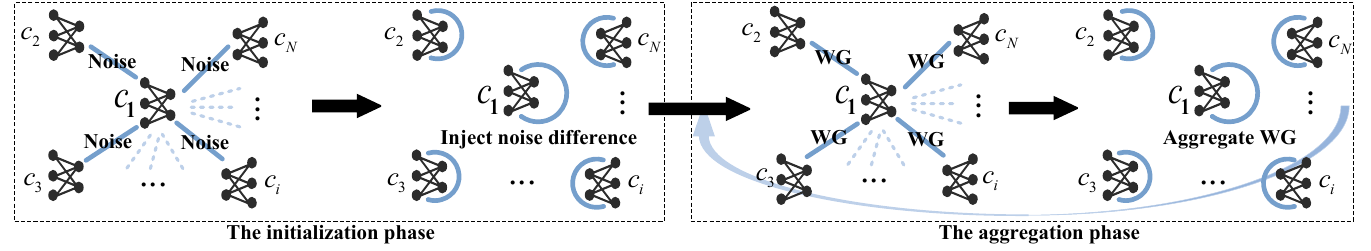}
    \vspace{-20pt}
 \caption{The procedure illustration of \textsf{LPPA} (from the  behaviors of client $c_1$), ``WG'' represents the model weight and gradient tracking variables for each client. Straight lines indicate information transmission and looped lines illustrate local aggregation.}
   \label{alg_1}
\end{figure*}


In this section, we first analyze the defect of the existing DP method for protecting gradients in the DSGT aggregation. We then propose our novel aggregation \textsf{LPPA} as a solution.



When it comes to enhancing gradient protection within DSGT aggregation, a \emph{straightforward} method is to add random noise into the local gradient tracking variable before transmission, which is exactly the same as the DP method \cite{10387777}. 
This typical DP method relies on the assumption of  ``internal isolation'' \cite{abadi2016deep}. It posits that, the adversary \emph{cannot access} other clients' local data and model, interacting \emph{only} with the received gradients. However, in the DSGT aggregation process (designed for DFL), all clients collaborate on global model training, the adversary can access critical information such as data dimension and model structure. 
It thus allows the adversary to reversely reconstruct sender's raw data based on the received gradient tracking variable.   
Adding noise into this transmitted variable only \emph{marginally} diminishes the integrity of reconstructed data, such protection is far from effective.  
In addition, increasing the added noise level in such a DP method will exacerbate the difficulty of consistent convergence across DFL clients. It then leads to insufficient gradient aggregation and significant degradation in global model accuracy.
Hence, even though much added noise benefits enhanced gradient protection, unlimited noise addition is impractical for high-accuracy consideration \cite{wei2020federated}.


Alternatively, in this paper, we propose a new solution, named \textsf{LPPA}, based on the standard DSGT aggregation rule. 
It adaptively utilizes the gradient tracking property in DSGT to enhance gradient protection.
\textsf{LPPA} further exploits the \emph{noise flow conservation}, where the overall noise inflow equals the noise outflow within the DFL connected topology (i.e., noise flows only within this topology), so as to achieve global noise elimination for lossless  model accuracy.

In particular, \textsf{LPPA} distinguishes the initialization from the subsequent aggregation. 
It just exchanges random noise in advance via an extra communication round, while it skillfully injects \emph{noise difference} into the initial gradient tracking variable before transmission.  
Compared to the pure noise addition like DP method, there are \emph{two key advantages} of noise difference injection in \textsf{LPPA}.
\emph{First}, for local gradient protection of each client, injecting noise difference between sent noise and received noise incorporates stronger randomness, as the adversary cannot deduce the noise levels of all sender's neighbors.
\emph{Second}, for lossless accuracy guarantee, noise flow conservation ensures that the accuracy of the global model aggregated by \textsf{LPPA} can be equivalent to that of standard DSGT.
Because the global sum of all noise differences remains zero per iteration and the gradient tracking variables primarily track the average of all local gradients, the noise differences \emph{do not affect this average value}. Hence, \textsf{LPPA} can guarantee accurate aggregation of gradient tracking variables and maintain lossless model accuracy.

Fig. \ref{alg_1} illustrates the general procedure of \textsf{LPPA} from the behaviors of the client $c_1$. 
It primarily consists of two phases. First, \emph{in the initialization phase}, each client sends random noise to each of its out-neighbors, and receives random noise from each of its in-neighbors. Then, each client calculates the noise difference between the total sent noise and the total received noise, and subsequently injects this noise difference into the local gradient tracking variable. 
Second, \emph{in the aggregation phase}, each client sends the model weight and injected gradient tracking variable to each of its out-neighbors. Upon receiving these variables from each of its in-neighbors, each client then aggregates these variables to update its own local model weight and gradient tracking variable for next transmission and aggregation. Generally, \textsf{LPPA} employs an extra communication round to exchange noise before transmitting model weights and gradient tracking variables. It then injects the noise difference for gradient protection and relies on the sum of all noise differences remaining zero to maintain lossless model accuracy.
\SetNlSty{normalfont}{\footnotesize}{:} 
\setlength{\textfloatsep}{5pt} 
\begin{algorithm}[t]
\small
\caption{\centering{The procedure of \textsf{LPPA}}}
\DontPrintSemicolon
\LinesNumbered
\SetNlSkip{0.4em}
\label{alg:dsgt} \KwIn{$N$ clients $C=\left\{c_1, c_2, \ldots, c_N\right\}$, the dataset $\mathcal{D}_i$, the out-neighbor clients $c_l \in \mathcal{N}_i^{out}$ and in-neighbor clients $c_j \in \mathcal{N}_i^{in}$ for each client $c_i$, the double-stochastic aggregation matrix $\boldsymbol{W}=[w_{ij}]_{n \times n}$, the decay step size $\lambda$ and the number of total communication rounds $T$\;}
\KwOut{The optimal and consistent global model weights $\boldsymbol{\theta}^{*}$}

\tcc{\textbf{Initialization}}
\For{\emph{each client} $c_i\in {C}$ \emph{in parallel}}{
set the local model weight $\boldsymbol{\theta}_i^0$ randomly\;
derive the initial local gradient $\nabla \mathcal{L}(\boldsymbol{\theta}_i^0)$ based on $\boldsymbol{\theta}_i^0$ and $\mathcal{D}_i$\;
send random noise $\boldsymbol{\delta}_{li}$ to each of its out-neighbors $c_l$\;
receive random noise $\boldsymbol{\delta}_{ij}$ from each of its in-neighbors $c_j$\;
 $\boldsymbol{\gamma}_i^0 \leftarrow \nabla \mathcal{L}(\boldsymbol{\theta}_i^0)+(\sum\nolimits_{c_l \in \mathcal{N}_i^{out}} \boldsymbol{\delta}_{li}-\sum\nolimits_{c_j \in \mathcal{N}_i^{in}} \boldsymbol{\delta}_{ij})$  \qquad \qquad \qquad \qquad
\tcp{{\scriptsize inject noise difference to local gradient tracking variable}}
}
\tcc{\textbf{Aggregation}}
\For{\emph{round} $t=0$ \emph{to} $(T-1)$}{ 
\For{\emph{each client} $c_i\in {C}$ \emph{in parallel}} {
send $\boldsymbol{\theta}_i^t$ and $\boldsymbol{\gamma}_i^t$ to each of its out-neighbors $c_l$\;
receive $\boldsymbol{\theta}_j^t$ and $\boldsymbol{\gamma}_j^t$ from each of its in-neighbors $c_j$ \;
 $\boldsymbol{\theta}_i^{t+1} \leftarrow \sum_{j=1}^N w_{ij}   \boldsymbol{\theta}_j^t -\lambda   \boldsymbol{\gamma}_i^t$ \qquad \qquad \qquad \qquad \qquad \qquad \qquad
\tcp{{\scriptsize aggregation of model weight}}
derive local gradient $\nabla \mathcal{L}(\boldsymbol{\theta}_i^t)$ based on $\boldsymbol{\theta}_i^t$ and $\mathcal{D}_i$\;
derive local gradient $\nabla \mathcal{L}(\boldsymbol{\theta}_i^{t+1})$ based on $\boldsymbol{\theta}_i^{t+1}$ and $\mathcal{D}_i$\;
 $\boldsymbol{\gamma}_i^{t+1} \leftarrow \sum_{j=1}^N w_{ij}   \boldsymbol{\gamma}_j^t + \nabla \mathcal{L}(\boldsymbol{\theta}_i^{t+1})-\nabla \mathcal{L}(\boldsymbol{\theta}_i^{t})$  \qquad \qquad \qquad \qquad 
\tcp{{\scriptsize aggregation of gradient tracking variable}}
}}
\Return $ \boldsymbol{\theta}_1^{T}= \cdots = \boldsymbol{\theta}_N^{T} = \boldsymbol{\theta}^{*}$ \;
\end{algorithm}

Algorithm \ref{alg:dsgt} provides the pseudo-code of \textsf{LPPA}. It takes $N$ clients $C=\left\{c_1, c_2, \ldots, c_N\right\}$, the dataset $\mathcal{D}_i$, the out-neighbor clients $c_l \in \mathcal{N}_i^{out}$ and in-neighbor clients $c_j \in \mathcal{N}_i^{in}$ for each client $c_i$,  the double-stochastic aggregation matrix $\boldsymbol{W}=[w_{ij}]_{n \times n}$ generated by common algorithms \cite{pu2020push}, the decay step size $\lambda$, and the number of total communication rounds $T$ as inputs. It outputs the optimal and consistent model weight $\boldsymbol{\theta}^{*}$, corresponding to the global model. 

At initialization, \textsf{LPPA} primarily injects noise difference into the local gradient tracking variable for gradient protection. Specifically, each client $c_i$ initially randomizes its local model weights $\boldsymbol{\theta}_i^0$ (line 2). Then, client $c_i$ derives the initial local gradient $\nabla \mathcal{L}(\boldsymbol{\theta}_i^0)$ based on its local dataset $\mathcal{D}_i$ and model weight $\boldsymbol{\theta}_i^0$ (line 3). Subsequently, client $c_i$ generates random noise $\boldsymbol{\delta}_{li}$ (i.e., $\boldsymbol{\delta}_{l \leftarrow i}$) and sends it to each of its out-neighbors $c_l$, $c_i$ also receives random noise $\boldsymbol{\delta}_{ij}$ from each of its in-neighbors $c_j$ (lines 4-5). Then $c_i$ calculates the noise difference $(\sum_{c_l \in \mathcal{N}_i^{out}} {\boldsymbol\delta}_{li}-\sum_{c_j \in \mathcal{N}_i^{in}} \boldsymbol{\delta}_{ij})$ by subtracting total noise sent from total noise received,  and injects this noise difference into its local gradient tracking variable $\boldsymbol{\gamma}_i^0$ for gradient protection (line 6).

In the subsequent aggregation phase, each client $c_i$ sends the model weight $\boldsymbol{\theta}^t_i$ and injected gradient tracking variable $\boldsymbol{\gamma}^t_i$ to each 
of its out-neighbors, and receives these variables from each of its in-neighbors in the $t$-th communication round (lines 9-10). Then they can aggregate the received model weights to update their own local model weights and derive the corresponding local gradients (lines 11-13). Subsequently, they aggregate received gradient tracking variables and update its gradient tracking variable for next transmission and aggregation (line 14). This process is repeated multiple times until the total number of communication rounds is reached. Eventually, \textsf{LPPA} returns the optimal and consistent model weight $ \boldsymbol{\theta}_1^{T}= \ldots = \boldsymbol{\theta}_N^{T} = \boldsymbol{\theta}^{*}$, corresponding to global model.

In conclusion, the aggregation form of \textsf{LPPA} for model weights and gradient tracking variables is the same as that of standard DSGT aggregation. \emph{The sole distinction} lies in the injection of noise difference into initial gradient tracking variables during the initialization phase (line 6). This noise difference injection is crucial, as it facilitates enhanced gradient protection while maintaining the global model's accuracy without any loss, as described in Section \ref{section-5}.
\begin{figure*}[!t]
    \centering
  \captionsetup[subfloat]{font=footnotesize} 
    \subfloat[The noise exchange among three clients]{
        \includegraphics[scale=0.50]{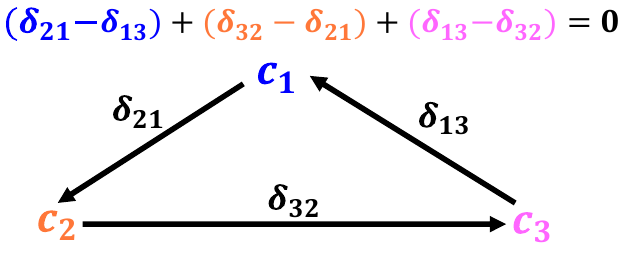}
        \label{guarantee_1}
    }
    \hspace{10pt}
    \subfloat[The matrix representation of noise exchange and further explanation]{
        \includegraphics[scale=0.50]{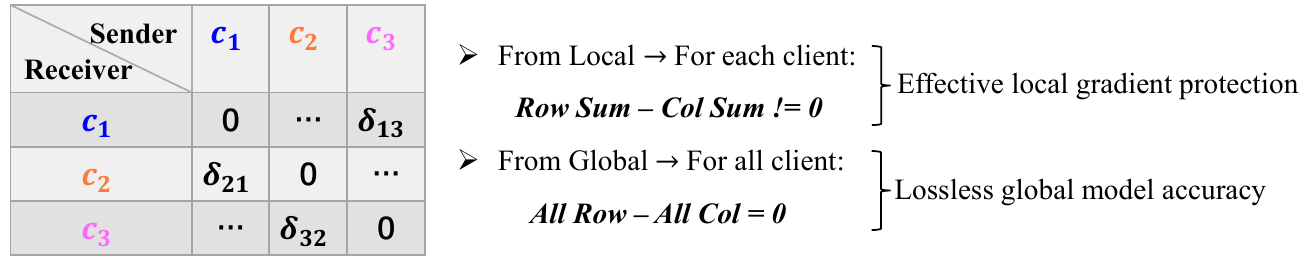}
        \label{guarantee_2}
    }
    \vspace{-7pt}
    \caption{The insight of the LPPA solution from both local and global perspectives.}
    \vspace{-10pt}
    \label{guarantee}
\end{figure*}

\section{Performance Analysis}
\label{section-5}

In this section, we examine the lossless global model accuracy derived by  \textsf{LPPA} relative to the standard DSGT aggregation (without noise injection). We also analyze its privacy-preserving capability relative to the DP method.

\subsection{\textbf{Lossless Global Model Accuracy}}
\label{section-5.1}



We first illustrate the noise flow
conservation in a simple DFL connected topology. Then, we prove the lossless global model accuracy under the proposed \textsf{LPPA} aggregation rule. 

Following, we present an example of noise exchange in a directed connected topology with three clients, as shown in Fig. \ref{guarantee_1}. In this simple topology, each client sends one random noise to its out-neighbor, and receives one random noise from its in-neighbor, for instance, $c_1$ sends $\boldsymbol{\delta}_{21}$ (i.e., $\boldsymbol{\delta}_{2 \leftarrow 1}$) to $c_2$ and receives $\boldsymbol{\delta}_{13}$ from $c_3$. Then, each client can calculate the noise difference, such as $(\boldsymbol{\delta}_{21}-\boldsymbol{\delta}_{13})$ for $c_1$. It is readily apparent that the global sum of all noise differences across three clients equals zero:
\begin{align}
\begin{aligned}
    \nonumber
    &(\boldsymbol{\delta}_{21}-\boldsymbol{\delta}_{13})+(\boldsymbol{\delta}_{32}-\boldsymbol{\delta}_{21})+(\boldsymbol{\delta}_{13}-\boldsymbol{\delta}_{32})\\
    =&(\boldsymbol{\delta}_{21}+\boldsymbol{\delta}_{32}+\boldsymbol{\delta}_{13})-(\boldsymbol{\delta}_{21}+\boldsymbol{\delta}_{32}+\boldsymbol{\delta}_{13})=\boldsymbol{0}
\end{aligned}
\end{align}
If the noise exchange is represented by a matrix, as shown in Fig. \ref{guarantee_2},  the noise difference injected by client $c_1$ actually corresponds to the difference between the \emph{$1$-th row} sum and the \emph{$1$-th column} sum in the matrix. 
This value is generally \emph{nonzero}.
Furthermore, from a global perspective, the sum of all noise differences is equivalent to the difference between \emph{all row} sums and \emph{all column} sums. This value is always \emph{zero}.
Generally, within the DFL connected topology, the global sum of all noise differences remains zero. Hence, the unaffected gradient aggregation and lossless accuracy can be guaranteed, as stated in Theorem \ref{lossless} (Proof in Appendix \ref{appendix_2}). 

\begin{theorem}
\label{lossless}
Given the DFL connected topology, all injected noise differences can be globally eliminated, then the accuracy of global model aggregated by \textsf{LPPA} can be equivalent to that of standard DSGT without noise injection. Namely, \textsf{LPPA} attains lossless accuracy.
\end{theorem}

\subsection{\textbf{Effective Local Gradient Protection}}
\label{section-5.2}
With the support of \textsf{LPPA},  the DFL aggregation not only harvests the lossless model accuracy but also puts more protection on local gradients and clients' raw data.  It is essential to explore the privacy-preserving capacity of  \textsf{LPPA} that utilizes noise difference injection.

In particular, we use the common concept of ``privacy budget'' in the DP method \cite{abadi2016deep} to gauge the privacy-preserving capacity of \textsf{LPPA}.
Let  $\Delta f$ represent the function sensitivity and  $\beta$ represent the Laplace noise scale parameter. This privacy budget  $\epsilon= \Delta f / \beta$ quantifies the extent of privacy loss, where smaller values of $\epsilon$ indicate a stronger privacy-preserving capacity.


\textbf{Privacy Budget in \textsf{LPPA}.} Assume that both  the function sensitivity vector $\boldsymbol{\Delta f}$ of DFL model and the Laplace noise scale parameter vector $\boldsymbol{\beta}$  across all clients in \textsf{LPPA} have been determined. Then we can deduce the privacy budget in \textsf{LPPA}, as outlined in Theorem \ref{noise_theo_1} (Proof in Appendix \ref{appendix_3}). At the $t$-th round, the privacy budget vector in \textsf{LPPA} is inversely proportional to \emph{$\sqrt{2}$ times} $\boldsymbol{\beta}$, and inversely proportional to the $t$-power aggregation matrix $\boldsymbol{W}^t$.

\begin{theorem}
\label{noise_theo_1}
Assume the function sensitivity vector in \textsf{LPPA} is denoted by $\boldsymbol{\Delta f}$ and the noise scale parameter vector across all clients in \textsf{LPPA} is denoted by $\boldsymbol{\beta}$.  The privacy budget vector of \textsf{LPPA} in the $t$-th round can then be expressed as $\boldsymbol{\Delta f} / \boldsymbol{W}^t \sqrt{2}\boldsymbol{\beta}$.
\end{theorem}


\textbf{Privacy Budget in DP.} In order to compare privacy-preserving capacity between \textsf{LPPA} and DP method, we also derive the privacy budget in the DP method under identical conditions. Because \textsf{LPPA} only injects the noise difference in the initialization and relies on this noise difference for gradient protection in subsequent aggregation. We establish a fair comparison by allowing the DP method to also add Laplace noise only in the initialization. Then we can derive the privacy budget in DP method is inversely proportional to scale parameter $\boldsymbol{\beta}$ and inversely proportional to the $t$-power aggregation matrix $\boldsymbol{W}^t$, as stated in Theorem \ref{noise_theo_2} (Proof in Appendix \ref{appendix_3}).

\begin{theorem}
\label{noise_theo_2}
Assume the function sensitivity vector in the DP method is represented by $\boldsymbol{\Delta f}$  and the noise scale parameter vector in the DP method is $\boldsymbol{\beta}$. The privacy budget vector of DP in the $t$-th round can be expressed as $\boldsymbol{\Delta f} / \boldsymbol{W}^t\boldsymbol{\beta}$.
\end{theorem}

\textbf{Comparison.} After getting the privacy budget in \textsf{LPPA} and DP method, we can simply deduce that the privacy-preserving capacity of \textsf{LPPA} is \emph{$\sqrt{2}$ times} that of DP method (i.e., the privacy budget of \textsf{LPPA} is \emph{$1/\sqrt{2}$ times} that of DP).
Furthermore, the DP method cannot add excessive noise level without diminishing the model accuracy, while \textsf{LPPA}’s global noise elimination allows for the injection of a more substantial noise difference level, thus further enhancing the gradient protection and data security. 

\emph{It is essential to emphasize that the noise difference can be injected in each communication round within the LPPA rule}. In our \textsf{LPPA} implementation, we confine the noise difference injection to the initialization phase to minimize communication overhead. Injecting noise difference or adding noise across multiple communication rounds could augment the privacy-preserving capabilities of the LPPA or DP method. Nevertheless, the \emph{distinctive advantage} of \textsf{LPPA} is that all noise differences can be globally eliminated, thereby not affecting the DFL aggregation process and ensuring lossless global model accuracy. In contrast, the DP method's noise injection typically results in a substantial accuracy degradation.

\begin{figure}[!t]
\vspace{-5pt}
    \centering
  \captionsetup[subfloat]{font=footnotesize} 
    \subfloat[The training curves]{
        \includegraphics[scale=0.28]{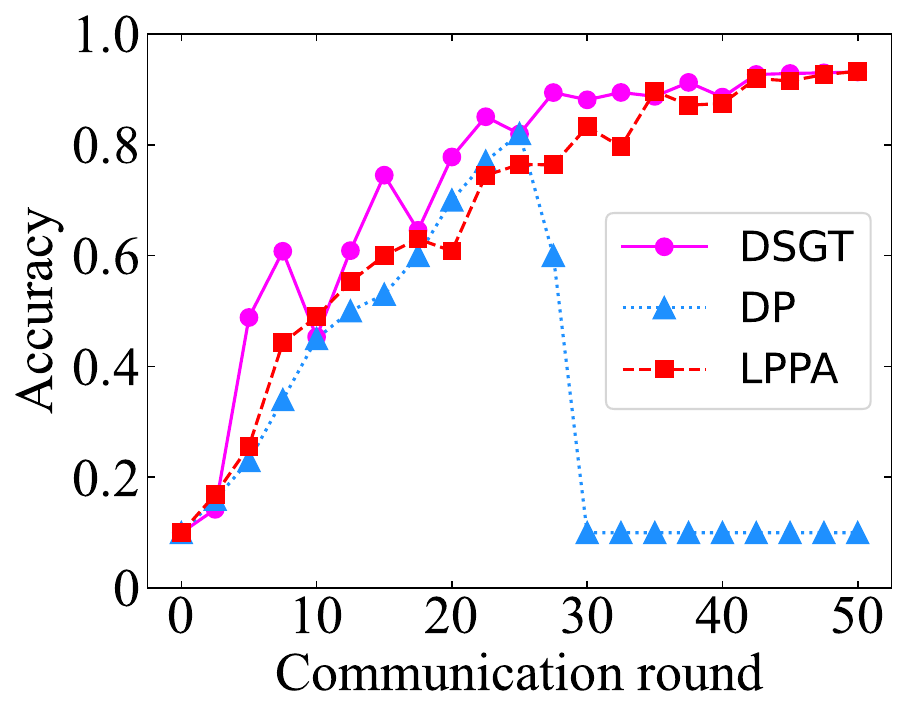}
        \label{fig:training-curve}
    }
    \subfloat[The DLG attack curves]{
    \hspace{-6pt}
        \includegraphics[scale=0.28]{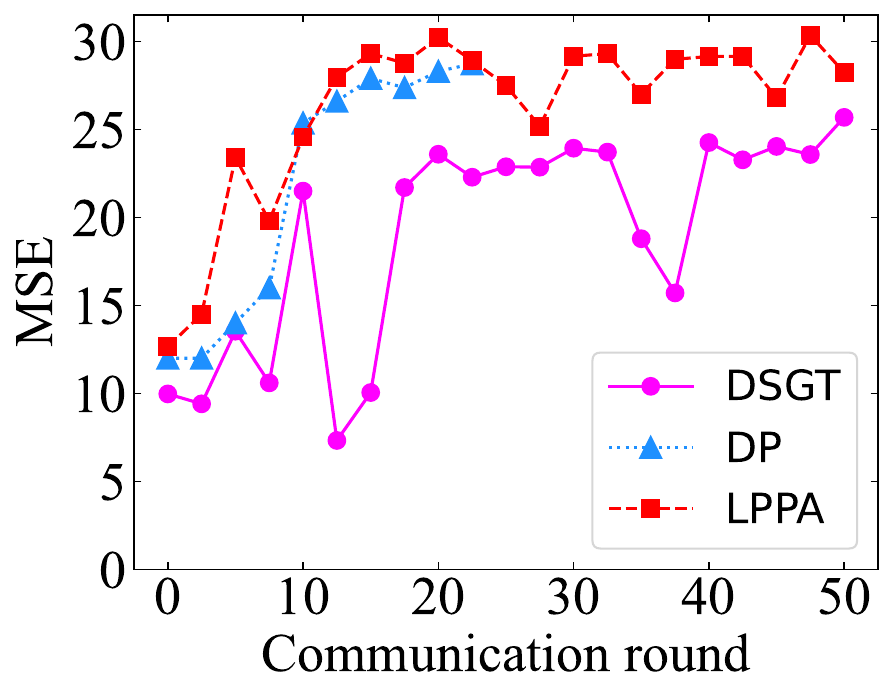}
        \label{fig:dlg-curve}
    }
    \vspace{-7pt}
    \caption{The performance vs. communication round.}
    \label{fig:mainfig-performance}
\end{figure}
\section{Experimental Evaluation}
\label{section-6}

In this section, we evaluate the performance of our proposed solution and other baselines. The experiments were conducted on an Intel Core 2.80GHz server with TITAN Xp 12GiB (GPU) and 192GB RAM, running the Ubuntu 18.04 system.



{\color{black}\textbf{Datasets.} In the experiments, we utilize three public real-world image datasets and three popular tabular datasets. For image classification, we use \textit{MNIST} \cite{deng2012mnist}, \textit{FMNIST} \cite{xiao2017fashion} and \textit{CIFAR-10} \cite{krizhevsky2009learning}. For binary classification, we use \textit{Adult} \cite{asuncion2007uci}, \textit{Rcv1} \cite{lewis2004rcv1} and \textit{Covtype} \cite{asuncion2007uci}.}

\textbf{Baselines.}  In the experiments, we evaluate three aggregation rules: the standard DSGT aggregation rule without noise injection (i.e., DSGT) for reference \cite{8618708}, DSGT aggregation rule within DP method \cite{10387777} (i.e., DP), and our proposed aggregation rule \textsf{LPPA}.

\textbf{Metrics.} 
In the evaluation, \emph{standard accuracy} (\emph{Accuracy}) and \emph{mean square error} (\emph{MSE}) are used to assess the model performance and privacy-preserving capability of \textsf{LPPA}, respectively. \emph{Accuracy} is defined as the likelihood that a well-trained global model correctly labels the input data in the testing dataset.  We also evaluate the accuracy loss (i.e., \emph{Loss}) between the other two rules and the standard DSGT, that is, the model accuracy derived by standard DSGT \emph{minus} the accuracy derived by \textsf{LPPA}/DP method. Moreover, \emph{MSE} measures the mean square error between the raw data and the reconstructed  data from the real data reconstruction attack DLG \cite{DBLP:series/lncs/Zhu020}.
Higher accuracy indicates better model performance. Similarly, higher MSE indicates greater privacy-preserving capacity. 
Each metric value is reported after averaging results five times.

\textbf{Implementation details.}
In the default experimental setup, five clients under full-connected communication topology were configured. The double-stochastic aggregation weight matrix is generated by the Sinkhorn-Knopp algorithm \cite{knight2008sinkhorn} and is fixed in one evaluation among three aggregation rules. The training dataset across the five clients is independently and identically distributed (i.e., homogeneous data partitioning), with the training dataset evenly divided into five equal parts for each client. The decay step size is by default set to 0.05, and the batch size is configured at 256. A total of 50 communication rounds is set to enable each aggregation rule to achieve a stable state, with the local epoch defaulting to 1. Laplace noise with a scale parameter of 0.025 is employed for noise generation. In particular, each client using the DP method adds $\boldsymbol{\beta=0.025}$  noise into the gradient tracking variable before transmission. Similarly, each client in \textsf{LPPA} exchanges $\boldsymbol{\beta=0.025}$  noise in advance via an extra communication round.
For the image classification, a two-layer convolutional neural network (CNN) is utilized as the global model structure. For the binary classification, a multi-layer perceptron (MLP) network is applied.

\begin{table}[!t]
\centering
\vspace{-7pt}
\caption{Accuracy under homogeneous data partitioning}
\vspace{-7pt}
\label{tab:accuracy_1}
\setlength{\tabcolsep}{8pt}
\begin{tabular}{|c|c|c|c|}
\hline
Dataset & DSGT & DP & \textsf{LPPA} \\ 
\hline
\textit{MNIST}  & $92.17 \pm 0.556$ & $90.87 \pm 0.801$ & $92.31 \pm 0.718$\\  \hline
\cline{1-1} \cline{2-4}
\textit{FMNIST} &   $74.25 \pm 0.666$ & $67.01 \pm 6.309$ & $74.63 \pm 0.888$\\ \hline
\cline{1-1} \cline{2-4}
\textit{CIFAR10} &   $39.24 \pm 0.755$ & $25.29 \pm 1.721$ & $39.33 \pm 0.764$\\ \hline
\cline{1-1} \cline{2-4}
\textit{Adult} &  $84.42 \pm 0.069$ & $75.70 \pm 0.013$ & $84.33 \pm 0.084$\\ \hline
\cline{1-1} \cline{2-4}
\textit{Rcv1} &   $91.03 \pm 0.229$ & $56.07 \pm 0.010$ & $91.03 \pm 0.263$\\ \hline
\cline{1-1} \cline{2-4}
\textit{Covtype} &  $76.62 \pm 0.168$ & $64.10 \pm 0.005$ & $76.62 \pm 0.132$\\
\hline
\textbf{Loss}  & {$-$} & $\boldsymbol{+13.12}$ & $\boldsymbol{-0.09}$ \\ \hline
\end{tabular}
\end{table}

\subsection{ \textbf{Accuracy Comparison}}
In the first set of experiments, we first plot the training curves for the standard DSGT, DP, and LPPA aggregation rules in Fig. \ref{fig:training-curve}. We also compare the \emph{highest} model accuracy of these rules under homogeneous data partitioning  (i.e., IID) in Table \ref{tab:accuracy_1}. 

Firstly, Fig. \ref{fig:training-curve} depicts the training curves of these rules over \textit{MNIST} datasets. It is evident that the accuracy of three rules improved initially, while \textsf{LPPA} and standard DSGT eventually achieved the equivalent accuracy. However, the accuracy of the DP method plummeted midway, thus preventing the completion of DFL aggregation. This aggregation failure is due to the irreducible noise added by the DP method, which disrupts the consistent convergence and eventually diminishes the model accuracy. In particular, in the early stages of DFL aggregation, when optimization is prioritized, the DP noise has minimal impact.
However, when aiming for consistency at a specific accuracy level in the aggregation process, all clients are unable to converge to consensus due to the irreducible noise in local gradients.
It thus amplifies the noise impact, leading to a ``NAN'' model weight value in the experiments and a sudden training failure.
In contrast, through noise flow conservation, the noise differences in \textsf{LPPA} can be globally eliminated to guarantee unaffected DFL aggregation and lossless model accuracy.

\begin{figure}[t]
\vspace{1pt}
\centering
            \vspace{-7pt}
 \captionsetup[subfloat]{font=footnotesize} 
	\subfloat[DSGT]{
\raggedleft
 \label{dlg.1}
\includegraphics[scale=0.99,trim=10 0 0 0]{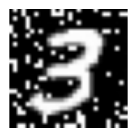}}
\hspace{0.20in}
 	\subfloat[DP]{
   \label{dlg.2}
			\includegraphics[scale=0.99]{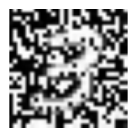}}
\hspace{0.20in}
 	\subfloat[\textsf{LPPA}]{
   \label{dlg.3}
   \raggedright
			\includegraphics[scale=0.99,trim=0 0 10 0]{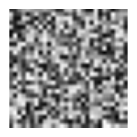}}
   \vspace{-10pt}
	\caption{\centering{The reconstructed data under DLG attack.}}
	\label{DLG.main}
\end{figure}
\begin{figure}[t]
    \centering
      \vspace{-15pt}
  \captionsetup[subfloat]{font=footnotesize} 
    \subfloat[The model accuracy]{
        \includegraphics[scale=0.27]{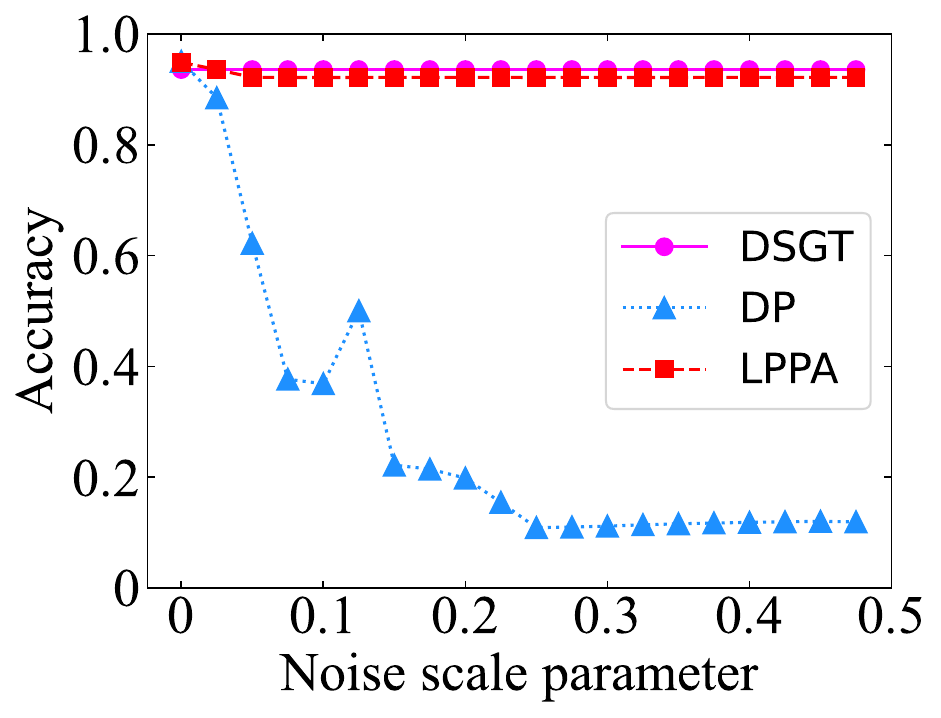}
        \label{fig:subfig1}
    }
    \subfloat[The privacy-preserving capacity]{
    \hspace{-9pt}
        \includegraphics[scale=0.27]{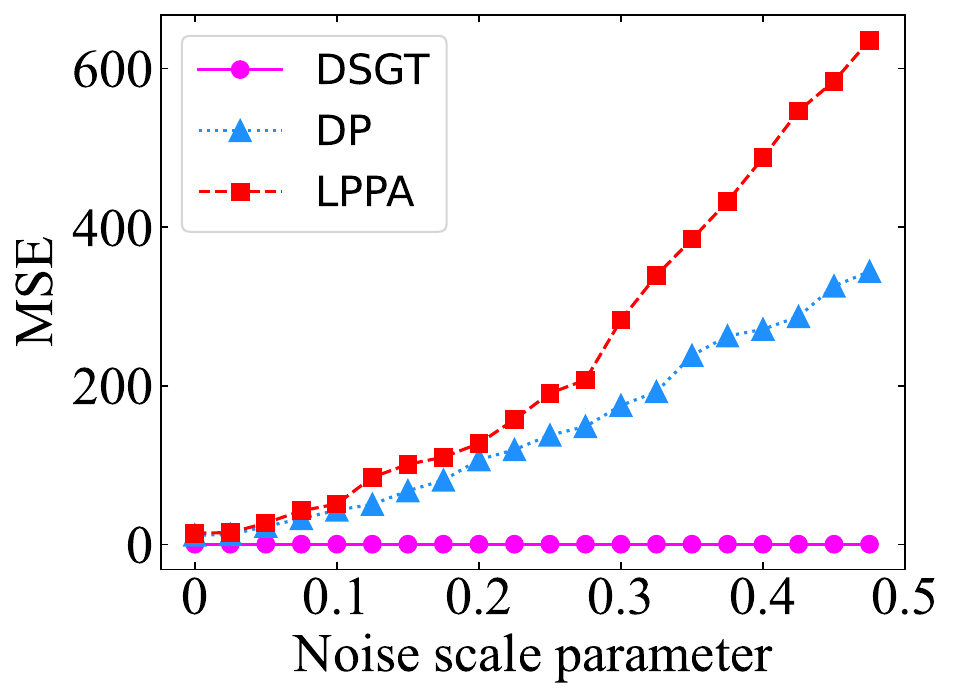}
        \label{fig:subfig2}
    }
    \vspace{-7pt}
    \caption{The performance vs. noise scale parameter.}
    \label{fig:mainfig-noise}
\end{figure}
As reported in Table \ref{tab:accuracy_1}, the model accuracy of these aggregation rules indicates that \textsf{LPPA} significantly outperforms the DP method.
Additionally, it achieves a global model accuracy approximately equivalent to that of standard DSGT, while even improving the model accuracy (i.e., the \emph{Loss is negative}).
Quantitatively, within the ideal homogeneous data partitioning, \textsf{LPPA} surpasses the DP method by a notable margin of 13.12\%, with this improvement escalating to 34.96\% and 14.04\% on the more complex \textit{Rcv1} and \textit{CIFAR10} datasets, respectively.
This marked superiority of \textsf{LPPA} can be ascribed to its robust capability to globally eliminate all noise differences, thereby ensuring accurate gradient aggregation and maintaining the lossless model accuracy.
Conversely, the noise added by the DP method is irreducible, inevitably encumbering the DFL aggregation and diminishing model accuracy, as stated in \cite{wei2020federated}.

\subsection{\textbf{Privacy-Preserving Capacity Comparison}}
In the second set of experiments, we use an actual data reconstruction attack DLG \cite{DBLP:series/lncs/Zhu020} to explicitly reveal the privacy-preserving capabilities of three aggregation rules.
This attack is a type of attack where an adversary exploits the gradients of a machine learning model to reconstruct the raw training data. \emph{MSE} represents the gap between reconstructed data and raw training data.
The larger value of MSE indicates a stronger privacy-preserving capacity.

Firstly, Fig. \ref{fig:dlg-curve} depicts the MSE curves of these rules under the actual DLG attack over \textit{MNIST} datasets. Due to noise interference, the model weight of the DP method midway reaches a ``NAN'' value, rendering the MSE incalculable (i.e., no plot in the second half of the figure). Nevertheless, even with the minimal noise scale parameter (i.e., 0.025), it is evident that the privacy-preserving capacity of \textsf{LPPA} surpasses that of DP method and is significantly superior to the standard DSGT. It is because the noise difference in \textsf{LPPA} can incorporate stronger randomness from various neighbors and introduce a larger noise scale parameter, as stated in Theorem \ref{noise_theo_1} and Theorem \ref{noise_theo_2}, thereby proving more effective gradient protection.

We also conduct the DLG attack over the image datasets \textit{MNIST} to visually demonstrate the feasibility and severity of data leaks. Fig. \ref{dlg.1} visualizes the reconstructed data from the DLG attack. One can observe that, the standard DSGT significantly compromises data privacy (i.e., complete reconstructed data). While the DP method provides marginally limited protection over standard DSGT, as shown in Fig. \ref{dlg.2}, close examination still reveals that the reconstructed image is the digit \emph{``3''}.
This finding further confirms that the noise added in the DP method only slightly diminishes the integrity of the reconstructed data, as the adversary in DFL \emph{already} retains sufficient information to reconstruct raw data.  In contrast, \textsf{LPPA} effectively prevents the adversary from reconstructing raw data, as shown in Fig. \ref{dlg.3}, underscoring its superior privacy-preserving capability again. This efficacy can be attributed to \textsf{LPPA}'s strategy of preemptively confusing clients through an extra round of noise exchange, which allows each client's gradient to blend with stronger randomness introduced by its various neighbors.

\begin{table}[t]
\centering
\vspace{-7pt}
\caption{MSE under DLG attack}
\vspace{-7pt}
\label{tab:privacy}
\setlength{\tabcolsep}{14pt}
\begin{tabular}{|c|S[table-format=4.3]|S[table-format=4.3]|S[table-format=4.3]|}
\hline
\multirow{2}{*}{Dataset} & \multicolumn{3}{c|}{MSE}\\ 
\cline{2-4} 
 & \multicolumn{1}{c|}{DSGT} & \multicolumn{1}{c|}{DP} & \multicolumn{1}{c|}{\textsf{LPPA}} \\ 
\hline
\textit{MNIST} & 0.372 & 9.828 & 12.039  \\ 
\hline
\textit{FMNIST} & 1.651 & 10.790 & 15.095 \\ 
\hline
\textit{CIFAR10} & 30.255 & 61.311 & 63.620 \\ 
\hline
\textit{Adult} & 0.057 & 2.606 & 3.355 \\ 
\hline
\textit{Rcv1} & 216.365 & 215.443 & 216.829 \\ 
\hline
\textit{Covtype} & 7897.451 & 7897.445 & 7897.598  \\ 
\hline
\end{tabular}
\end{table}
\begin{figure}[t]
    \centering
    \vspace{-15pt}
  \captionsetup[subfloat]{font=footnotesize} 
    \subfloat[Different local epoch]{
        \includegraphics[scale=0.27]{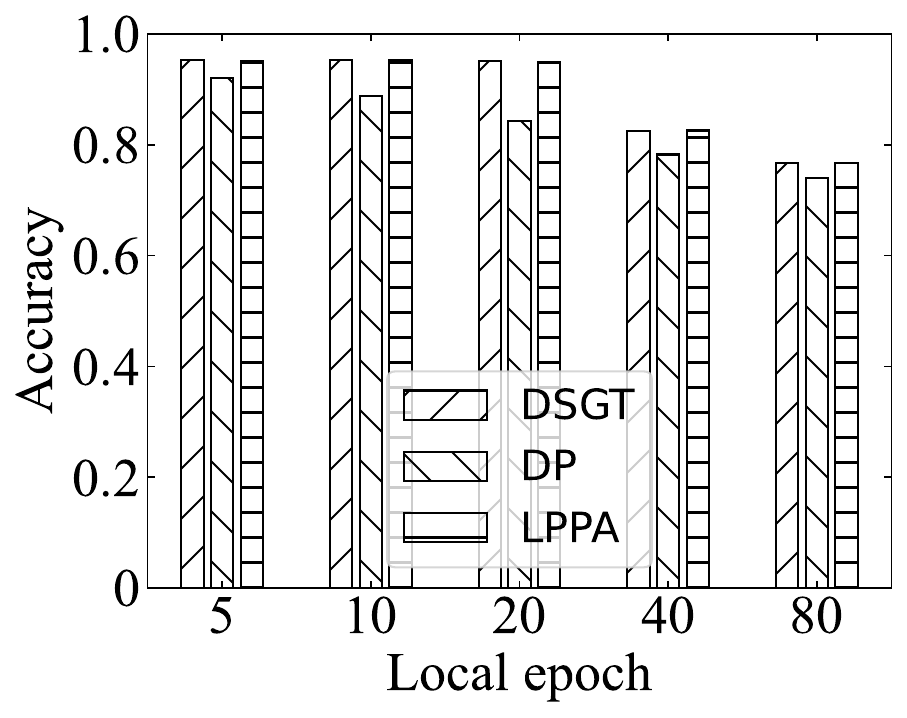}
        \label{fig:local-epoch}
    }
    \subfloat[Different number of clients]{
        \includegraphics[scale=0.27]{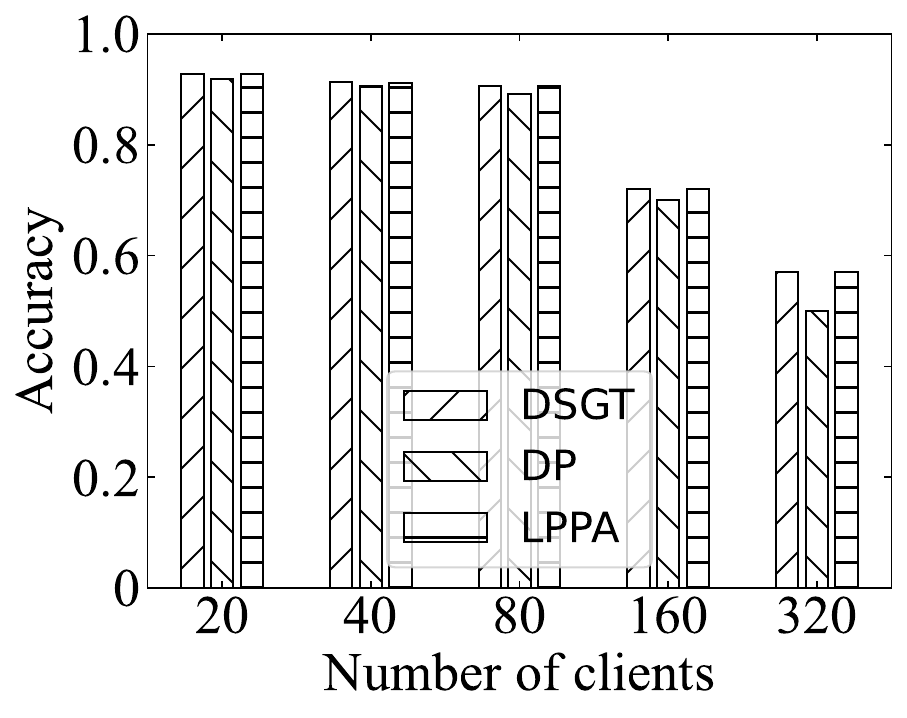}
        \label{fig:clients}
    }
    \vspace{-7pt}
    \caption{The performance vs. different hyperparameters.}
    \label{fig:mainfig-1}
\end{figure}

Table \ref{tab:privacy} reports the specific privacy-preserving capacity of these rules over each dataset. 
The variance in MSE across different datasets is primarily attributed to differences in data dimensions. Notably, the \textit{Rcv1} and \textit{Covtype} datasets, which feature higher data dimensions, exhibit increased MSE due to the enhanced complexity of data reconstruction.
Nevertheless, we can still find that, \textsf{LPPA} significantly outperforms both the DP method and standard DSGT.
Generally, \textsf{LPPA} surpasses standard DSGT without noise injection about \emph{tenfold} MSE improvement. Compared to the DP method, \textsf{LPPA} achieves about \emph{1.5 times} average improvement, which is consistent with the analysis in Theorem \ref{noise_theo_1}. 
The superior privacy-preserving capacity of \textsf{LPPA} is because that, the noise difference in \textsf{LPPA} enables a larger noise scale parameter than the DP method, thus providing more  effective gradient protection and powerful data security.




\subsection{\textbf{Parameter Evaluation}}
\emph{Effect of $\boldsymbol{\beta}$.}
When varying the noise scale parameter $\boldsymbol{\beta}$ from $\boldsymbol{0.025}$ to $\boldsymbol{0.5}$, the corresponding experimental results over the \textit{MNIST} dataset are presented in Fig. \ref{fig:mainfig-noise}. Note that the noise scale parameter $\boldsymbol{\beta}$ does not impact the performance of the standard DSGT.  \textsf{LPPA} consistently demonstrates superior privacy-preserving capability (i.e., MSE) and higher accuracy than the DP method in all cases. \textsf{LPPA} also achieves the accuracy \emph{equivalent} to that of standard DSGT and significantly \emph{enhances} the privacy-preserving capacity. Additionally, as $\boldsymbol{\beta}$ increases, the privacy-preserving capability of \textsf{LPPA} and DP method consistently improve, with \textsf{LPPA}'s capability approximately \emph{$\sqrt{2}$ times} that of DP method. However, the accuracy of DP gradually decreases, while the model accuracy of \textsf{LPPA} remains stable, still equivalent to that of standard DSGT. This is because the noise difference in \textsf{LPPA} can provide a larger scale parameter than the noise added in DP method, and all noise differences can be globally eliminated to ensure lossless global model accuracy.

\begin{table}[t]
\centering
\caption{Accuracy under quantity distribution skew}
\vspace{-7pt}
\label{tab:accuracy_3}
\setlength{\tabcolsep}{8pt}
\begin{tabular}{|c|c|c|c|}
\hline
Dataset & DSGT & DP & \textsf{LPPA} \\ 
\hline
\textit{MNIST}   & $91.34 \pm 0.889$& $90.62 \pm 1.006$& $91.60 \pm 0.710$\\ \hline
\cline{1-1} \cline{2-4}
\textit{FMNIST}  & $73.94 \pm 0.791$& $65.49 \pm 7.730$& $73.73 \pm 0.847$\\ \hline
\cline{1-1} \cline{2-4}
\textit{CIFAR10} & $39.12 \pm 0.757$& $26.63 \pm 0.377$& $39.60 \pm 1.175$\\ \hline
\cline{1-1} \cline{2-4}
\textit{Adult} & $83.62 \pm 0.685$& $75.70 \pm 0.016$& $83.48 \pm 0.650$\\ \hline
\cline{1-1} \cline{2-4}
\textit{Rcv1} & $90.65 \pm 0.257$& $53.68 \pm 4.872$& $90.61 \pm 0.347$\\ \hline
\cline{1-1} \cline{2-4}
\textit{Covtype} & $76.19 \pm 0.648$& $64.10 \pm 0.013$& $75.94 \pm 0.885$\\
\hline
\textbf{Loss}  & {$-$} &  $\boldsymbol{+13.27}$ & $\boldsymbol{-0.02}$ \\ \hline
\end{tabular}
\end{table}

\emph{Effect of local epoch.} Upon varying the local epoch (i.e., the number of local updates performed before sending model weights and gradient tracking variables to neighbors) from 5 to 80, the corresponding experimental results (i.e., accuracy) over the \textit{MNIST} dataset are depicted in Fig. \ref{fig:local-epoch}. It is observed that, as the local epoch increases, the model accuracy of each aggregation rule consistently decreases. This is because, as the local epoch increases, the clients in DFL cannot obtain enough information from neighboring clients on time. It is more likely to result in local optima, thus hindering productive aggregation. Moreover, \textsf{LPPA} achieves higher model accuracy than the DP method. It also attains equivalent accuracy to that of standard DSGT. It is because the irreducible noise added by DP amplifies the potential for convergence to local optima and causes training failure. In contrast, the robustness of \textsf{LPPA} is attributed to its timely global noise elimination.

\emph{Effect of clients' number.}
We examine the effect of the number of clients on these aggregation rules, as shown in Fig. \ref{fig:clients}. It is observed that the accuracy decreases consistently with an increase in the number of clients. The reason is that, when the number is large, the amount of local dataset per client is reduced. It thus potentially leads to overfitting during local training and a decrease in the model accuracy. In this case, the accuracy of \textsf{LPPA} still performs better than DP method and can be competitive with the standard DSGT.

\begin{table}[t]
\centering
\caption{Accuracy under label distribution skew}
\vspace{-7pt}
\label{tab:accuracy}
\setlength{\tabcolsep}{1.3pt}
\begin{tabular}{|c|c|c|c|c|}
\hline
Dataset & Partitioning & DSGT & DP & \textsf{LPPA} \\ 
\hline
    

  \multirow{2}{*}{\textit{MNIST}} & $p_k \sim \text{Dir}(0.1)$ 
  & $91.32 \pm 0.803$& $88.25 \pm 4.121$& $91.15 \pm 0.628$ \\
  \cline{2-5}
  & $\#C=2$  & $90.98 \pm 0.552$& $88.67 \pm 1.990$& $91.27 \pm 0.647$\\ \hline
 
  \multirow{2}{*}{\textit{FMNIST}} & $p_k \sim \text{Dir}(0.1)$ 
  & $72.02 \pm 0.143$& $44.88 \pm 6.801$& $72.04 \pm 0.566$\\
\cline{2-5}
  & $\#C=2$ 
  & $72.23 \pm 0.559$& $49.97 \pm 6.510$& $72.58 \pm 0.186$\\ 
  \hline

  \multirow{2}{*}{\textit{CIFAR10}} & $p_k \sim \text{Dir}(0.1)$ 
  
  &$33.11 \pm 0.330$& $11.97 \pm 2.236$& $32.34 \pm 0.360$\\ 
\cline{2-5}
  & $\#C=2$ & $33.54 \pm 1.049$& $17.10 \pm 2.769$& $32.94 \pm 0.095$\\ \hline

  \multirow{2}{*}{\textit{Adult}} & $p_k \sim \text{Dir}(0.1)$ & $81.00 \pm 2.908$& $75.66 \pm 0.082$& $79.54 \pm 2.850$\\
\cline{2-5}
  & $\#C=1$ & $82.23 \pm 0.210$& $75.70 \pm 0.004$& $82.08 \pm 0.359$\\ \hline

 \multirow{2}{*}{\textit{Rcv1}} & $p_k \sim \text{Dir}(0.1)$ & $90.98 \pm 0.186$& $50.86 \pm 4.955$ &$90.73 \pm 0.300$\\
\cline{2-5}
  & $\#C=1$ & $91.08 \pm 0.216$& $56.07 \pm 0.006$& $91.18 \pm 0.249$\\ \hline

 \multirow{2}{*}{\textit{Covtype}} & $p_k \sim \text{Dir}(0.1)$ & $72.32 \pm 2.405$& $64.10 \pm 0.001$& $73.02 \pm 4.605$\\
\cline{2-5}
  & $\#C=1$ & $76.35 \pm 0.130$& $64.10 \pm 0.004$& $76.30 \pm 0.140$\\
 \hline
\multicolumn{2}{|c|}{\textbf{Loss}}  & {$-$} & $\boldsymbol{+16.82}$ & $\boldsymbol{+0.19}$ \\ 
\hline 
\end{tabular}
\end{table}

\subsection{\textbf{Data Partitioning Evaluation}}
In the last set of experiments, we evaluate the robustness and universality of these aggregation rules under more realistic scenarios. We adopt two most common \emph{heterogeneous} data partitioning strategies \cite{li2022federated}: quantity distribution skew and label distribution skew, to simulate the real-world scenarios.
Note that, we do not use the partitioning of feature distribution skew where diverse noise is added to the raw data. Because our proposed LPPA and DP both need to add noise difference/noise,  we avoid this partition in order to avoid possible interference. 
In the quantity distribution skew, the distribution of the number of data categories among the five clients follows the Dirichlet distribution \cite{DBLP:journals/bioinformatics/WuSWCC17}.
Under the label distribution skew, the label distribution of clients is also divided according to the Dirichlet distribution  (e.g.,  $p_k \sim \text{Dir}(0.1)$), or each client holds only a few labels (e.g., $\#C=2$, each client only has two labels).

Within the realistic heterogeneous data partitioning, our proposed \textsf{LPPA} surpasses DP method by an impressive margin of 13.29\% (resp. 16.63\%), achieving peak improvements of 12.97\% (resp. 20.37\%) on the \textit{CIFAR10} dataset under quantity (resp. label) distribution skew, as shown as in Table \ref{tab:accuracy_3} and Table \ref{tab:accuracy}. In comparison to the standard DSGT, \textsf{LPPA} exhibits a negligible margin in accuracy under quantity distribution skew and label distribution skew. All observed variations in accuracy between \textsf{LPPA} and standard DSGT fall within the accepted margin of random error. Furthermore, \textsf{LPPA} demonstrates stable model accuracy and strong robustness under heterogeneous data partitioning. This stability is attributed to \textsf{LPPA}'s ability to globally eliminate the noise impact by summing all noise differences to zero. It thereby facilitates stable and accurate gradient aggregation among all clients. Conversely, the irreducible DP noise exacerbates the disparity in clients' local gradient under heterogeneous data partitioning. It thus leads to serious aggregation failures and a huge decline in global model accuracy.

\vspace*{-6pt}
\section{Conclusion}
\label{section-7}
In this paper, we propose a novel privacy-preserving aggregation rule  \textsf{LPPA} for decentralized federated learning. It
remarkably improves the privacy-preserving capability of the standard DFL aggregation rule without noise injection, while flawlessly maintaining the lossless global model accuracy.  
It subtly injects the noise difference between sent noise and received noise into local gradients to enhance gradient protection. 
The global sum of all noise differences being zero enables \textsf{LPPA} harvest lossless model accuracy eventually. 
We theoretically confirm the lossless global model accuracy and analyze the superior privacy-preserving capacity of \textsf{LPPA}.
Extensive experiments over six real-world datasets demonstrate that, \textsf{LPPA} significantly outperforms the state-of-the-art methods in both model accuracy and privacy preservation. 

\bibliographystyle{ACM-Reference-Format}
\balance
\bibliography{lppa}

\clearpage
\setcounter{section}{0}
\renewcommand{\thesection}{\Alph{section}}
\section{Summary of Notations}
\begin{table}[ht]
    \centering
     \caption{The description of the symbols}
    \label{tab:symbol_label}
\resizebox{0.48\textwidth}{!}{%
\Large
\begin{tabular}{|c|l|}
\hline
\textbf{Symbol} & \textbf{Meaning} \\ \hline
$N$ & the number of clients\\ \hline
$c_i \in C$ & the set of clients\\ \hline
$\mathcal{D}_i$ & the local dataset for client $c_i$\\ \hline
$\mathcal{D}$ & the collective union of all local datasets\\ \hline
$c_l \in \mathcal{N}_i^{out}$ &  each of out-neighbors for client $c_i$ (receive information from $c_i$)\\ \hline
$c_j \in \mathcal{N}_i^{in}$ & each of in-neighbors for client $c_i$ (send information to $c_i$)\\ \hline
$w_{ij}$ &the aggregation weight assigned by $c_i$ for its in-neighbor $c_j$\\ \hline
$\boldsymbol{W}$ &the aggregation weight matrix unified by aggregation weight $w_{ij}$ \\ \hline
$\mathcal{L}(\cdot)$ &the loss function across all clients\\ \hline
$T$ &the number of total communication rounds \\ \hline
$\boldsymbol{\theta}_i^t$ &the local model weight for client $c_i$ in the $t$-th  communication round \\ \hline
$\nabla \mathcal{L}(\boldsymbol{\theta}_i^t)$ &the local gradient for client $c_i$ in the $t$-th communication round \\ \hline
$\boldsymbol{\gamma}_i^t$ &the local gradient tracking variable for client $c_i$ in the $t$-th  round \\ \hline
$\boldsymbol{\delta}_{li}$& the random noise sent by $c_i$ to its out-neighbor $c_l$\\ \hline
$\boldsymbol{\delta}_{ij}$ &the random noise received by $c_i$ from its in-neighbor $c_j$\\ \hline
$\lambda$ &the decay step size \\ \hline
$\boldsymbol{\theta}^{*}$ &the optimal and consistent model weight corresponding to global model\\ \hline
$\beta$  &The scale parameter of Laplace noise\\ \hline
$\Delta f$ & The function sensitivity \\ \hline
\end{tabular}}
\end{table}

\section{Proof of Theorem \ref{lossless}}
\label{appendix_2}
\begin{proof}

The predictive accuracy of DFL global model essentially depends on the convergence dynamic of DFL aggregation. Therefore, to prove the Theorem \ref{lossless}, we mainly confirm that the accurate aggregation of gradient tracking variable and model weight within \textsf{LPPA} remain unaffected, compared with the standard DSGT.

It can be noted that the aggregation form of \textsf{LPPA} for gradient tracking variables is the same as that of standard DSGT. Hence, we firstly prove the unaffected global gradient tracking property and then state the unaffected gradient  aggregation in \textsf{LPPA}.

Since $\sum_{i=1}^N w_{ij}=1$ in the double-stochastic aggregation weight matrix $\boldsymbol{W}$, the aggregation of gradient tracking variable can be added up as follows.
\begin{align}
	\begin{aligned}
 \nonumber
	\sum_{i=1}^N \boldsymbol{\gamma}_i^{t+1}&=\sum_{i=1}^N \sum_{j=1}^N w_{ij}  \boldsymbol{\gamma}_j^t+\sum_{i=1}^{N} (\nabla \mathcal{L}(\boldsymbol{\theta}^{t+1}_i)-\nabla \mathcal{L}(\boldsymbol{\theta}^{t}_i))\\
	&=\sum_{j=1}^N \boldsymbol{\gamma}_j^t   \sum_{i=1}^N w_{ij}+\sum_{i=1}^N(\nabla \mathcal{L}(\boldsymbol{\theta}^{t+1}_i)-\nabla \mathcal{L}(\boldsymbol{\theta}^{t}_i))\\
	&=\sum_{i=1}^N \boldsymbol{\gamma}_i^t+\sum_{i=1}^N(\nabla \mathcal{L}(\boldsymbol{\theta}^{t+1}_i)-\nabla \mathcal{L}(\boldsymbol{\theta}^{t}_i))\\
\end{aligned}
\end{align}
It can be further rewritten as follows. 
\begin{align}
\begin{aligned}
 \nonumber
\sum_{i=1}^N (\boldsymbol{\gamma}_i^{t+1}-\nabla\mathcal{L}(\boldsymbol{\theta}^{t+1}_i)) 
=\sum_{i=1}^N(\boldsymbol{\gamma}_i^{t}-\nabla\mathcal{L}(\boldsymbol{\theta}^{t}_i))=\dots=\sum_{i=1}^N(\boldsymbol{\gamma}_i^0-\nabla\mathcal{L}(\boldsymbol{\theta}^0_i))
\end{aligned}
\end{align}

In the initialization phase of \textsf{LPPA}, each initial gradient tracking variable $\boldsymbol{\gamma}_i^0$ is injected into the noise difference. 
\begin{align}
\begin{aligned}
    \nonumber
          \boldsymbol{\gamma}^{0}_i= \nabla \mathcal{L}(\boldsymbol{\theta}_i^0)+
          (\sum\nolimits_{c_l \in \mathcal{N}_i^{out}} \boldsymbol{\delta}_{li}-\sum\nolimits_{c_j \in \mathcal{N}_i^{in}} \boldsymbol{\delta}_{ij})
\end{aligned}
\end{align}




Without loss of generality, the random noise between each two clients that are not communicating can be marked as 0. Then the global sum of all noise differences can be derived to equal zero, i.e.,
\begin{align}
\nonumber
\begin{aligned}
\sum_{i=1}^{N}(\sum\nolimits_{c_l \in \mathcal{N}_i^{out}} \boldsymbol{\delta}_{li}-\sum\nolimits_{c_j \in \mathcal{N}_i^{in}} \boldsymbol{\delta}_{ij})= \sum_{i=1}^{N}\sum_{l=1}^{N} \boldsymbol{\delta}_{li}-\sum_{i=1}^{N}\sum_{j=1}^{N} \boldsymbol{\delta}_{ij}=\boldsymbol{0}
\end{aligned}  
\end{align}
Due to the global sum
of all noise differences being equal to zero, we can deduce $\sum_{i=1}^N \boldsymbol{\gamma}_i^0=\sum_{i=1}^N \nabla \mathcal{L}(\boldsymbol{\theta}_i^0)$ as follows. 
\begin{align}
	\begin{aligned}
 \nonumber
		\sum_{i=1}^N \boldsymbol{\gamma}_i^{0}
		&= \nabla \mathcal{L}(\boldsymbol{\theta}_i^0)+
          (\sum\nolimits_{c_l \in \mathcal{N}_i^{out}} \boldsymbol{\delta}_{li}-\sum\nolimits_{c_j \in \mathcal{N}_i^{in}} \boldsymbol{\delta}_{ij}) \\
          &=\sum_{i=1}^N   \nabla \mathcal{L}(\boldsymbol{\theta}_i^0)+\boldsymbol{0} =\sum_{i=1}^N   \nabla \mathcal{L}(\boldsymbol{\theta}_i^0)
	\end{aligned}
\end{align}
Thus, we can affirm that $\sum_{i=1}^N (\boldsymbol{\gamma}_i^{0} - \nabla \mathcal{L}(\boldsymbol{\theta}_i^0)) =\boldsymbol{0}$, which consequently guarantees that $\sum_{i=1}^N \boldsymbol{\gamma}_i^t = \sum_{i=1}^N \nabla \mathcal{L}(\boldsymbol{\theta}_i^t)$ in \textsf{LPPA}. It then deduce the \emph{perfect gradient tracking property and unaffected aggregation of gradient tracking variable} in \textsf{LPPA}.

Regarding the aggregation of model weight in \textsf{LPPA}, after noise difference injection, it also can be added up as follows.
\begin{align}
\nonumber
\begin{aligned}
 \sum_{i=1}^N \boldsymbol{\theta}_i^{1}=& \sum_{i=1}^N \sum_{j=1}^N w_{ij}   \boldsymbol{\theta}_j^0 -\lambda \sum_{i=1}^N  \boldsymbol{\gamma}^0_i\\
 =& \sum_{i=1}^N \sum_{j=1}^N w_{ij}   \boldsymbol{\theta}_j^0 -\lambda \sum_{i=1}^N\nabla \mathcal{L}(\boldsymbol{\theta}_i^0)\\
 &+
         \sum_{i=1}^N( \sum\nolimits_{c_l \in \mathcal{N}_i^{out}} \boldsymbol{\delta}_{li}-\sum\nolimits_{c_j \in \mathcal{N}_i^{in}} \boldsymbol{\delta}_{ij}) \\
 =&\sum_{i=1}^N \sum_{j=1}^N w_{ij}   \boldsymbol{\theta}_j^0 -\lambda \sum_{i=1}^N \nabla \mathcal{L}(\boldsymbol{\theta}_i^0)
\end{aligned}
\end{align}
Hence, the aggregation of model weight remains unaffected in the initialization.
The model weight in the subsequent aggregation process is updated by the unaffected gradient tracking variable, thus the aggregation of model weight can also remain unaffected.
Therefore, \textsf{LPPA} can ensures accurate DFL aggregation of model weight and gradient tracking variable, thereby harvesting the lossless model accuracy.
\end{proof}

\section{Proof of Theorem \ref{noise_theo_1} and Theorem \ref{noise_theo_2}}
\label{appendix_3}

It should be noted in advance tha, we use the common concept of ``privacy budget'' in the DP method \cite{abadi2016deep} to gauge the privacy-preserving capacity of \textsf{LPPA}.
Before defining the privacy budget, we need to introduce the related concepts of \emph{function sensitivity} and \emph{Laplace noise scale parameter}.
Specifically, function sensitivity, represented by $\Delta f$, measures the maximal impact that any individual client's data sample can have on its local gradient \cite{abadi2016deep}.
Based on the function sensitivity of the DFL model, the DP method typically employs mechanisms such as the Laplace mechanism, which adds Laplace noise into local gradients.
The Laplace noise scale parameter, denoted as $\beta$, which governs the noise magnitude, is pivotal. It ensures that the added noise effectively masks the influence of any individual client's data, adhering to the established threshold by privacy budget $\epsilon$:
\begin{align}
    \begin{aligned}
        \nonumber
       \epsilon= \Delta f / \beta
    \end{aligned}
\end{align}
The privacy budget quantifies the extent of privacy loss, where smaller values of $\epsilon$ indicate a stronger privacy-preserving capacity.

\textit{Notion.} Following the concept of privacy budget in the DP method, we can deduce the privacy budget in \textsf{LPPA}. For ease of explanation, we need to make some notations. In particular, let $\boldsymbol{\Gamma}^t = [\boldsymbol{\gamma}^t_1, \ldots, \boldsymbol{\gamma}^t_N]$ represent the gradient tracking matrix unified by all gradient tracking variables in the $t$-th ($1 \leq t \leq T$) communication round of \textsf{LPPA}, and let ${\dot{\boldsymbol{\Gamma}}}^t$ represent the corresponding matrix of the standard DSGT. $\boldsymbol{\Gamma}_{DP}^t$ represents the corresponding matrix in DP method.
$\boldsymbol{W}^t$ denotes the $t$-th power of the double-stochastic aggregation weight matrix $\boldsymbol{W}$. 
Furthermore, $\boldsymbol{\delta}^S_i = \sum_{c_l \in \mathcal{N}_i^{out}} \boldsymbol{\delta}_{li}$ represents the total noise sent to out-neighbors for client $c_i$ in \textsf{LPPA}, and $\boldsymbol{\Delta}^S$ represents the noise matrix $[\boldsymbol{\delta}^S_1, \ldots, \boldsymbol{\delta}^S_N]$. Similarly, $\boldsymbol{\delta}^R_i = \sum_{c_j \in \mathcal{N}_i^{out}} \boldsymbol{\delta}_{ij}$ represents the total noise received from in-neighbors for client $c_i$ in \textsf{LPPA}, and $\boldsymbol{\Delta}^R$ represents the noise matrix $[\boldsymbol{\delta}^R_1, \ldots, \boldsymbol{\delta}^R_N]$. We use $\boldsymbol{\zeta}_i$ to denote the random noise added to the initial gradient tracking variable for client $c_i$ in the DP method, $\boldsymbol{Z}$ represents the noise matrix $[\boldsymbol{\zeta}_1, \ldots, \boldsymbol{\zeta}_N]$.
$\nabla \mathcal{L}(\boldsymbol{\Theta}^t)=[\nabla \mathcal{L}(\boldsymbol{\theta}_i^t), \ldots, \nabla \mathcal{L}(\boldsymbol{\theta}_N^t)]$ represents the matrix of local gradients in $t$-th communication round. 
Furthermore, $\beta_i$ represents the noise scale parameter of the Laplace noise applied by client $c_i$ and $\boldsymbol{\beta}$ represents the noise scale parameter vector across all clients. The function sensitivity vector across all clients in DFL is denoted by $\boldsymbol{\Delta f}$. 
Now, we are ready to inspect \emph{the privacy budget} in \textsf{LPPA}, and \emph{quantify} the privacy-preserving capacity per communication round in comparison to the DP method.



\begin{proof}
To demonstrate the Theorem \ref{noise_theo_1}, we first prove the \emph{noise perturbation} for gradient protection in \textsf{LPPA} at $t$-th communication round  can be expressed as $\boldsymbol{W}^t(\boldsymbol{\Delta}^S - \boldsymbol{\Delta}^R)$ and then deduce the privacy budget of \textsf{LPPA} is $\boldsymbol{\Delta f} / \sqrt{2}\boldsymbol{W}^t\boldsymbol{\beta}$. Note $\boldsymbol{W}^t \in \mathbb{R}^{N \times N}$ and $N$ is the total number of all DFL clients, $(\boldsymbol{\Delta}^S - \boldsymbol{\Delta}^R) \in \mathbb{R}^{N \times P}$ and $P$ is the dimension of model weight/gradient vector, $\boldsymbol{\Delta f}\in \mathbb{R}^{N \times 1}$ and $\boldsymbol{\beta}\in \mathbb{R}^{N \times 1}$ are function sensitivity vector and noise scale parameter vector across all clients.

We prove this noise perturbation using \emph{mathematical induction}. First, we establish that the claim holds for round $t=0$; then we assume its validity for round $t=k$ and demonstrate that the claim also holds for round $t=k+1$. 

\emph{In the base case}, we consider the round $t=0$ of \textsf{LPPA}. The standard gradient tracking matrix $\dot{\boldsymbol{\Gamma}}^0$, without noise injection, is given by $\dot{\boldsymbol{\Gamma}}^0 = \nabla \mathcal{L}(\boldsymbol{\Theta}^0)$. Thus, the gradient tracking variable matrix $\boldsymbol{\Gamma}^0$ in \textsf{LPPA} satisfies the stipulated conclusion, i.e.,
\begin{align}
\nonumber
\boldsymbol{\Gamma}^0 =  \nabla \mathcal{L}(\boldsymbol{\Theta}^0) + (\boldsymbol{\Delta}^S - \boldsymbol{\Delta}^R) = \dot{\boldsymbol{\Gamma}}^0 + \boldsymbol{W}^{0} (\boldsymbol{\Delta}^S - \boldsymbol{\Delta}^R),
\end{align}
where $\boldsymbol{\Delta}^S$ and $\boldsymbol{\Delta}^R$ are the matrix unified by each client's total noise sent and total noise received, respectively. 

\emph{In the inductive hypothesis}, we assume that the gradient tracking variable in the $t=k$ round satisfies the conclusion:
\begin{align}
\nonumber
\boldsymbol{\Gamma}^k = \dot{\boldsymbol{\Gamma}}^k + \boldsymbol{W}^k (\boldsymbol{\Delta}^S - \boldsymbol{\Delta}^R).
\end{align}

\emph{In the inductive step}, we need to prove this statement is true for $t=k+1$ round.
We can apply the aggregation rule of gradient tracking variable, i.e., project $\boldsymbol{\Gamma}^k$ into $\boldsymbol{\Gamma}^{k+1}$:

\begin{align}
\begin{aligned}
    \nonumber
\boldsymbol{\Gamma}^{k+1}&= \boldsymbol{W}   \boldsymbol{{\Gamma}}^k  + \nabla \mathcal{L}(\boldsymbol{\Theta}^{k+1})- \nabla \mathcal{L}(\boldsymbol{\Theta}^k) \\
&= \boldsymbol{W}   (\dot{\boldsymbol{\Gamma}}^k + \boldsymbol{W}^k (\boldsymbol{\Delta}^S - \boldsymbol{\Delta}^R)) +\nabla \mathcal{L}(\boldsymbol{\Theta}^{k+1})- \nabla \mathcal{L}(\boldsymbol{\Theta}^k) \\
&= \boldsymbol{W}   \dot{\boldsymbol{\Gamma}}^k + \nabla \mathcal{L}(\boldsymbol{\Theta}^{k+1})- \nabla \mathcal{L}(\boldsymbol{\Theta}^k) + \boldsymbol{W}   \cdot \boldsymbol{W}^k  (\boldsymbol{\Delta}^S - \boldsymbol{\Delta}^R) \\
&= \dot{\boldsymbol{\Gamma}}^{k+1} + \boldsymbol{W}^{k+1}  (\boldsymbol{\Delta}^S - \boldsymbol{\Delta}^R)
\end{aligned}
\end{align}
where $\dot{\boldsymbol{\Gamma}}^{k+1}$ represents the gradient tracking vector in the $t=k+1$  round of the standard DSGT. From the assumption in the $t=k$ round, the statement in the $t=k+1$ round can be derived; coupled with the validation for the $t=0$ round, then the noise perturbation for gradient protection in \textsf{LPPA} can be derived.

After getting the noise perturbation is $\boldsymbol{W}^t  (\boldsymbol{\Delta}^S - \boldsymbol{\Delta}^R)$, the scale parameter of Laplace noise $\boldsymbol{\Delta}^S$/$\boldsymbol{\Delta}^R$ is $\boldsymbol{\beta}$. Then we can know the scale parameter of $(\boldsymbol{\Delta}^S - \boldsymbol{\Delta}^R)$ is $\sqrt{2}\boldsymbol{\beta}$. And the function sensitivity vecotr in \textsf{LPPA} is $\boldsymbol{\Delta f}$, then we can deduce the privacy budget of \textsf{LPPA} according to \cite{abadi2016deep} is $\boldsymbol{\Delta f} / \sqrt{2}\boldsymbol{W}^t\boldsymbol{\beta}$. Note that this division is element-by-element division.

Similarly, to demonstrate the Theorem \ref{noise_theo_2}, we first prove the specific noise perturbation for clients in DP method at $t$-th round is expressed as $\boldsymbol{W}^t \boldsymbol{Z}$ and then deduce the privacy budget is $\boldsymbol{\Delta f}/\boldsymbol{W}^t\boldsymbol{\beta}$. We also use mathematical induction to prove it.


\emph{In the base case}, in the $t=0$ round,  the standard gradient tracking matrix $\dot{\boldsymbol{\Gamma}}^0$, devoid of noise injection, is given by $\dot{\boldsymbol{\Gamma}}^0 = \nabla \mathcal{L}(\boldsymbol{\Theta}^0)$. Then it can be deduced that the gradient tracking variable $\boldsymbol{\Gamma}_{DP}^0$ would fulfill the stipulated conclusion, i.e.,
\begin{align}
\nonumber
   \boldsymbol{\Gamma}_{DP}^0 = \nabla \mathcal{L}(\boldsymbol{\Theta}^0) + \boldsymbol{Z} = \dot{\boldsymbol{\Gamma}}^0 + \boldsymbol{W}^{0} \boldsymbol{Z}
\end{align}

\emph{In the inductive hypothesis}, we assume that the gradient tracking variable in the $k$-th round satisfies the conclusion:
\begin{align}
\nonumber
 \boldsymbol{\Gamma}_{DP}^k = \dot{\boldsymbol{\Gamma}}^k + \boldsymbol{W}^k \boldsymbol{Z}.
\end{align}

\emph{In the inductive step}, we need to prove the statement is true for $t=k+1$ round, and we project $\boldsymbol{\Gamma}_{DP}^k$ into $\boldsymbol{\Gamma}_{DP}^{k+1}$ through the aggregation rule, i.e.,
\begin{align}
\begin{aligned}
    \nonumber
 \boldsymbol{\Gamma}_{DP}^{k+1} &= \boldsymbol{W}   \boldsymbol{\Gamma}_{DP}^{k} + \nabla \mathcal{L}(\boldsymbol{\Theta}^{k+1})- \nabla \mathcal{L}(\boldsymbol{\Theta}^k)  \\
 &= \boldsymbol{W}   (\dot{\boldsymbol{\Gamma}}^k + \boldsymbol{W}^k \boldsymbol{Z}) + \nabla \mathcal{L}(\boldsymbol{\Theta}^{k+1})- \nabla \mathcal{L}(\boldsymbol{\Theta}^k)   \\
&= \boldsymbol{W}   \dot{\boldsymbol{\Gamma}}^k + \nabla \mathcal{L}(\boldsymbol{\Theta}^{k+1})- \nabla \mathcal{L}(\boldsymbol{\Theta}^k)   + \boldsymbol{W}   \cdot \boldsymbol{W}^k  \boldsymbol{Z} \\
&= \dot{\boldsymbol{\Gamma}}^{k+1} + \boldsymbol{W}^{k+1}  \boldsymbol{Z}
\end{aligned}
\end{align}
where $\dot{\boldsymbol{\Gamma}}^{k+1}$ represents the gradient tracking vector in the $t=k+1$ round of the standard DSGT. From the assumption in the $t=k$ round, the statement in the $t=k+1$ round can be derived; coupled with the validation for the $t=0$ round, then the noise perturbation for gradient protection in DP method can be derived.

After getting the noise perturbation is $\boldsymbol{W}^t  \boldsymbol{Z}$, the scale parameter vector of added Laplace noise is $\boldsymbol{\beta}$, and function sensitivity vector is $\boldsymbol{\Delta f}$, then we can deduce the privacy budget of DP method is $\boldsymbol{\Delta f}/\boldsymbol{W}^t \boldsymbol{\beta}$.
Note that this division is element-by-element division
\end{proof}


\end{document}